\pdfoutput=1
\documentclass[letterpaper, 10 pt, conference]{ieeeconf}

\IEEEoverridecommandlockouts
\usepackage{footmisc}
\usepackage{graphicx}
\DeclareGraphicsExtensions{.pdf,.png,.jpg,.jpeg}
\usepackage{amsmath,amssymb}
\usepackage{multirow}
\usepackage{hyperref}
\usepackage{listings}
\usepackage{xcolor}
\usepackage{booktabs}
\usepackage{subfig}

\title{\LARGE \bf
MERGE: Guided Vision‑Language Models for Multi‑Actor Event Reasoning and Grounding in Human–Robot Interaction
}

\author{Joerg Deigmoeller$^{*1}$, 
Nakul Agarwal$^{*2}$,
Stephan Hasler$^{1}$, 
Daniel Tanneberg$^{1}$, 
Anna Belardinelli$^{1}$, \\ 
Reza Ghoddoosian$^{2}$,
Chao Wang$^{1}$, 
Felix Ocker$^{1}$,
Fan Zhang$^{1}$, 
Behzad Dariush$^{2}$, 
Michael Gienger$^{1}$\\ 
\small $^{*}$equal contribution  
\thanks{$^{1}$Honda Research Institute Europe, 63073 Offenbach, Germany}
\thanks{$^{2}$Honda Research Institute USA, San Jose, CA 95134, USA.}
}

\begin{document}

\maketitle
\thispagestyle{empty}
\pagestyle{empty}

\begin{abstract}

We introduce MERGE, a system for situational grounding of actors, objects, and events in dynamic human–robot group interactions. Effective collaboration in such settings requires consistent situational awareness, built on persistent representations of people and objects and an episodic abstraction of events. MERGE achieves this by uniquely identifying physical instances of actors (humans or robots) and objects and structuring them into actor–action–object relations, ensuring temporal consistency across interactions.
Central to MERGE is the integration of Vision-Language Models (VLMs) guided with a perception pipeline: a lightweight streaming module continuously processes visual input to detect changes and selectively invokes the VLM only when necessary. This decoupled design preserves the reasoning power and zero-shot generalization of VLMs while improving efficiency, avoiding both the high monetary cost and the latency of frame-by-frame captioning that leads to fragmented and delayed outputs.
To address the absence of suitable benchmarks for multi-actor collaboration, we introduce the GROUND dataset, which offers fine-grained situational annotations of multi-person and human–robot interactions. On this dataset, our approach improves the average grounding score by a factor of 2 compared to the performance of VLM-only baselines--including GPT-4o, GPT-5 and Gemini 2.5 Flash--while also reducing run-time by a factor of 4. The code and data are available at \url{www.github.com/HRI-EU/merge}.


\end{abstract}

\section{INTRODUCTION}
\label{introduction}

Situational awareness in collaborative human–robot group environments is a multifaceted challenge; robots must continuously track who is doing what, where, and with whom as interactions evolve. Relying on isolated, micro-level observations of actions is insufficient; comprehensive understanding demands modeling individuals, their inter-relationships, and the broader group context simultaneously~\cite{jahangard2024jrdb}. For example, a robotic assistant in a team meeting or kitchen must maintain identities across view changes and capture evolving actor–action–object relationships (e.g., person A hands an object to person B) rather than just a set of unconnected detections. Achieving this level of situational grounding -- akin to structured “who-does-what-to-whom” recognition -- remains a fundamental hurdle for current vision systems in multi-actor settings.

Recent advances in large-scale VLMs and multimodal foundation models show strong visual reasoning, with systems such as GPT-4~\cite{achiam2023gpt} and PaLM-E~\cite{driess2023palme} demonstrating impressive general vision-language capabilities and newer frameworks like Gemini~\cite{team2024gemini}, LLaVA~\cite{li2024llava}, and Flamingo~\cite{alayrac2022flamingo} extending multimodal reasoning across images, audio, and video. However, these models primarily excel at class-level reasoning and do not inherently maintain instance-level identity over time. Applying them frame-by-frame to videos is computationally prohibitive for robotics and contextually inconsistent, since they process frames independently. This leads to failures in multi-actor settings -- for example, referring to “a person” or “the tool” in each frame without realizing it is the same entity. Moreover, recent evaluations show that state-of-the-art video VLMs over-rely on single-frame cues and struggle when only temporal relationships carry information, in some cases collapsing to near-zero accuracy~\cite{upadhyay2025time}. These limitations highlight the need for approaches that can selectively guide VLM reasoning with persistent, instance-aware, temporal grounding in dynamic multi-actor environments.


To address these gaps, we introduce MERGE, a system for multi-actor event
reasoning and grounding in human-robot group interactions. MERGE integrates a perception pipeline to \emph{guide} the VLM that uniquely identifies physical instances of persons and objects and structures their interactions as actor--action--object relations with temporal consistency. A lightweight streaming module continuously tracks actors and objects and detects salient changes; only then is the VLM invoked to render semantic judgments. This decoupled design preserves the reasoning power and zero-shot generalization of VLMs while ensuring efficiency, avoiding the prohibitive costs and fragmented outputs of frame-by-frame captioning, and obviating fine-tuning on small task-specific datasets.

Another obstacle is the lack of benchmarks for multi-actor situational grounding. Existing datasets largely focus on single-actor activities or lack the fine-grained, role-aware annotations required for collaborative HRI (Human-Robot Interaction). As a result, they do not provide sufficient support for studying persistent multi-actor grounding. To address this gap, we introduce GROUND\footnotemark[\value{footnote}], a dataset specifically designed for multi-actor human–robot interactions with detailed actor–action–object relations. GROUND enables systematic evaluation of situational grounding, role distinction, and multi-actor awareness in group interactions. Using GROUND, we evaluate MERGE on collaborative pouring, handovers, and sorting, showing that it reliably maintains multi-actor awareness, distinguishes roles, and generates significantly more reliable actor--action--object relationships across time over vanilla VLMs. Together, MERGE and GROUND provide a structured and efficient foundation for spatiotemporal reasoning and situated decision-making in human--robot collaboration, moving toward group-aware interactive robots.

In summary, the contributions of this work are:

\begin{itemize}
\item We introduce MERGE, a guided VLM framework that is VLM-independent, combining lightweight perception with selective VLM invocation to maintain persistent actor–object identities and generate structured event tuple efficiently.
\item We provide GROUND, a benchmark dataset of multi-human–robot collaborations with detailed annotations for role-aware situational grounding of interactions, and design new evaluation metrics.
\item We demonstrate that MERGE outperforms state-of-the-art methods in both accuracy and runtime, establishing a foundation for group-aware HRI.
\end{itemize}

\section{RELATED WORK}
\label{related_work}
\noindent \textbf{Grounding in Robotics: VLMs and Symbolic Frameworks.}
As robotics technology advances, integrating LLMs and VLMs has become essential for building autonomous systems capable of complex human interaction. Recent surveys \cite{zhang2023large, li2025benchmark} document the growing use of multi-modal foundation models to enhance robotic perception, reasoning, and decision-making. However, enabling seamless collaboration with humans still requires deep semantic context understanding and precise grounding of physical instances within the environment.
One research direction targets high-level instruction following. Systems such as Do-As-I-Can~\cite{brohan2023can}, Inner Monologue~\cite{huang2022inner}, and Mobile-ALOHA~\cite{fu2024mobile} ground language into executable actions, while vision-language-action models like RT-2~\cite{zitkovich2023rt} and OpenVLA~\cite{kim2024openvla} extend this paradigm by transferring web-scale knowledge into robotic execution. Despite their strengths, these approaches primarily address single-actor tasks and overlook collaborative settings.
A second line of work investigates video-based VLMs for general understanding. Models including R3M \cite{nair2022r3m}, Bringing Robots Home~\cite{shafiullah2023bringing}, VILA \cite{lin2024vila}, VideoLLaMA2 \cite{cheng2024videollama}, and Qwen-VL \cite{Qwen-VL} leverage multimodal inputs, including video, for tasks such as question answering and captioning. Frameworks like VideoAgent \cite{wang2024videoagent}, VideoTree \cite{wang2025videotree}, and MindPalace \cite{huang2025building} further pre-structure scene elements to improve long-video reasoning. While advancing large-scale video understanding, these methods do not address persistent, role-aware grounding for collaborative HRI.

Closer to our setting are robotics-specific efforts that combine grounding and reasoning. VLM-See-Robot-Do \cite{wang2024vlm} demonstrates how frame-wise analysis with object detection can improve grounding over raw video-based VLMs \cite{team2024gemini, li2024llava, lin2024vila}, yet remains restricted to single-actor demonstrations. Other approaches such as the Pyramid Graph Convolutional Network for spatio-temporal HOI \cite{xing2022understanding}, Robotic Visual Instruction \cite{li2025robotic}, Hi Robot \cite{shi2025hi}, and ManipLVM-R1 \cite{song2025maniplvm} extend vision-language(-action) models toward relational reasoning, visual instruction following, hierarchical task execution, and embodied reasoning with reinforcement learning. Despite these advances, they do not provide role-aware multi-actor grounding of collaborative interactions.

Finally, earlier symbolic frameworks explored complementary perspectives. Lemaignan et al. \cite{lemaignan2011you} grounded natural language through symbolic reasoning and perspective-taking, enabling robots to interpret vague situated expressions, but only in dyadic settings. The Object-Action Complex framework \cite{kruger2011object} provided a formalism for hierarchically organizing sensorimotor experiences into symbolic action representations, but without explicit modeling of human interaction. More recently, LaMI \cite{wang2024lami} incorporated LLMs for multi-modal reasoning in robot-group interaction, while the Attentive Support framework \cite{tanneberg2024help} introduced proactive assistance strategies for human groups. However, both approaches rely on marker-based object detection and lack scalability in open-world settings.

In summary, prior work has advanced grounding at the level of single-actor instruction following, general video-based perception, robot-centric grounding and reasoning, and symbolic or group interaction frameworks. Yet none of these approaches provide persistent, role-aware multi-actor grounding of interactions in collaborative human–robot scenarios. This gap motivates our work, which explicitly targets grounding of actors, objects, and their relations in dynamic group settings.

\noindent \textbf{Datasets for Grounding and HRI.} Several benchmarks across diverse domains have been introduced to support activity recognition and video understanding, including EPIC-KITCHENS \cite{damen2018scaling}, Ego4D \cite{grauman2022ego4d}, HD-EPIC \cite{damen2018scaling} and Rank2Tell~\cite{sachdeva2024rank2tell}. While these datasets provide rich annotations for everyday activities, they primarily capture single-actor or egocentric scenarios and thus fall short for studying collaborative human–robot interactions. JRDB-Social \cite{jahangard2024jrdb} emphasizes multi-person group dynamics, but it does not capture fine-grained manipulation or role-aware action tuples. Synthetic environments such as BEHAVIOR-1K \cite{li2023behavior} offer broad coverage of activities, yet lack ecological validity for real-world HRI. More recent large-scale evaluation benchmarks, such as Video-MME \cite{fu2025video} and EgoSchema \cite{mangalam2023egoschema}, focus on testing general video understanding or commonsense reasoning over narratives, often through multiple-choice question answering. While valuable for probing the high-level reasoning abilities of multimodal models, they lack the granularity required for fine-grained action reasoning and do not provide role-aware annotations of collaborative interactions. To the best of our knowledge, no publicly available dataset offers detailed situational annotations involving multiple interacting actors; the most closely related effort \cite{wang2024vlm} is not publicly available and does not address multi-actor scenarios.

\section{MERGE FRAMEWORK}
\label{frame_work}
\begin{figure*}[thpb]
  \centering
  \includegraphics[width=0.90\linewidth, height=6.0cm]{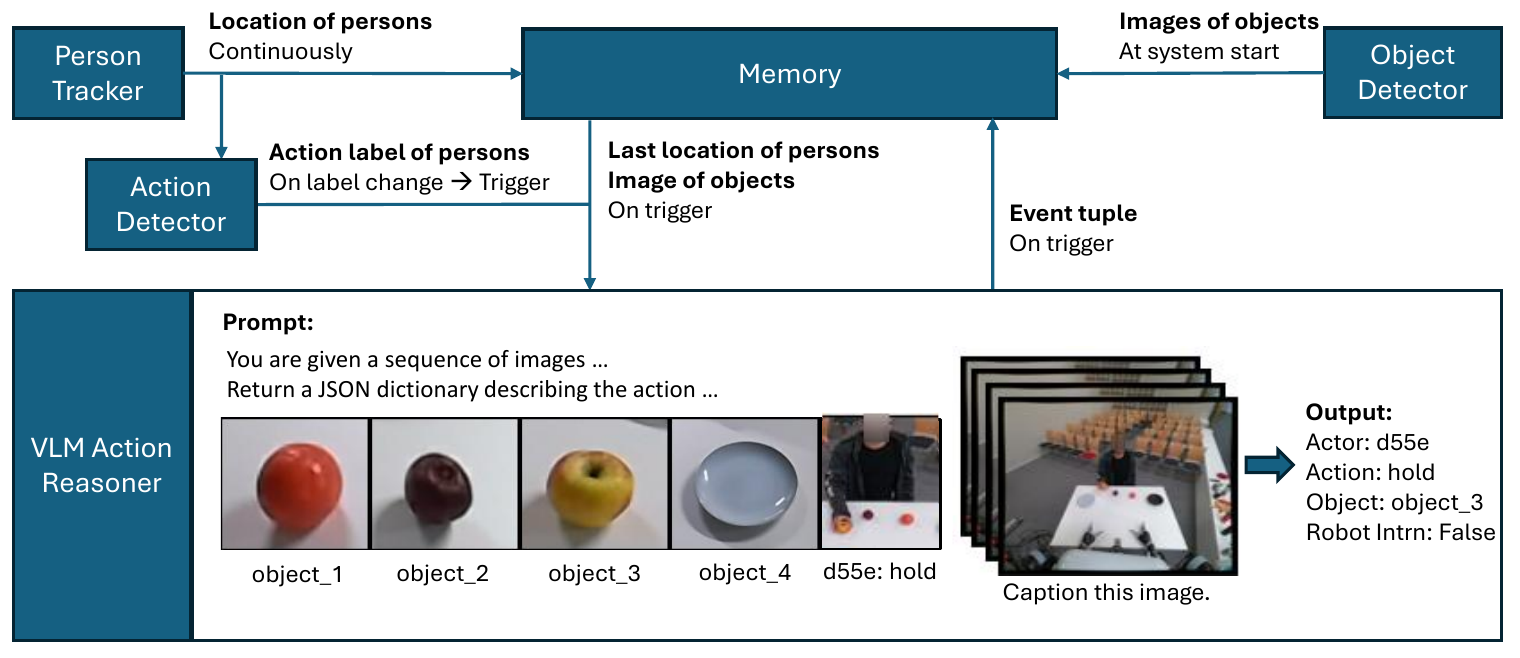}
  \caption{
  \textbf{Overview of MERGE}. The \textit{Action Reasoner} is a VLM that integrates three structured inputs: a region of the current camera frame centered on person\_y to capture the relevant scene entities and their spatial arrangement; reference images of previously detected objects\_x to ground reasoning in known instances; and the four recent images captured by image\_i. These inputs are sourced from the \textit{Memory} module, which aggregates person locations from the \textit{Person Tracker} and object references from the \textit{Object Detector} at system initialization. In parallel, a lightweight \textit{Action Detector} continuously predicts action labels from the image stream, and any change in a person’s predicted action triggers the Action Reasoner, with the updated label itself serving as an additional VLM input.}
  \label{figure:MERGE}
  \vspace{-10pt}
\end{figure*}

We introduce the MERGE (\textbf{M}ulti-actor \textbf{E}vent
\textbf{R}easoning and \textbf{G}rounding in Human–Robot Int\textbf{E}raction) framework by beginning with its output: a uniquely defined sequence of 
event tuples, denoted as $\mathcal{T} = {T_1, T_2, \ldots, T_n}$. Each event tuple $T = (a, x, o, r, t, i)$ encodes the fundamental elements of an event -- namely, \textit{who} ($a\in A$) performs \textit{what} action ($x\in X$) on \textit{which} object ($o\in O$), as well as \textit{where} ($r\in \mathcal{R}_{A,O} \cup \{\varnothing\}$), \textit{when} ($t\in \mathbb{R}_{\geq 0}$), and \textit{whether} the robot is involved ($i\in I$).
By organizing real-world events into such event tuple, the framework keeps track of distinct actors and objects in a way that remains consistent over time, even as people move around or objects get picked and placed.

To illustrate the functionality of the MERGE framework, consider a collaborative tabletop task where two persons and one robot engage in a shared interaction. One human actor ($a_1$) hands over an apple ($o_1$) to the robot ($a_3$). The robot then places the apple into a bowl ($o_2$), while the second human actor ($a_2$) picks up an orange ($o_3$) and places it also into the bowl ($o_2$). These actions are not independent but part of a coordinated fruit sorting task, where roles shift dynamically, and actions depend on prior handovers and object placements. Using the above notation, the scenario can be illustrated in temporal order with $t_1<t_2<t_3$ as follows:

\[
\mathcal{T} = 
\left\{
\begin{aligned}
T_1 &= (a_1,\ \text{hand over},\ o_1,\ (\text{to},\ a_3),\ t_1) \\
T_2 &= (a_3,\ \text{place down},\ o_1,\ (\text{in},\ o_2),\ t_2) \\
T_3 &= (a_2,\ \text{place down},\ o_3,\ (\text{in},\ o_2),\ t_3)
\end{aligned}
\right\}
\]

This structured event representation effectively serves as an episodic memory of the observed activities in a unified format. In practice, such a memory enables downstream AI reasoning modules (e.g. VLMs) to interpret group behavior with higher quality and consistency~\cite{huet2025episodic}. Moreover, by preserving temporal order and context, the MERGE representation provides a solid basis for reasoning about underlying causality – an AI can analyze the chronologically ordered events to identify possible cause-effect relationships.
\begin{figure*}[t]
  \centering
  \includegraphics[width=0.95\linewidth, height=10cm]{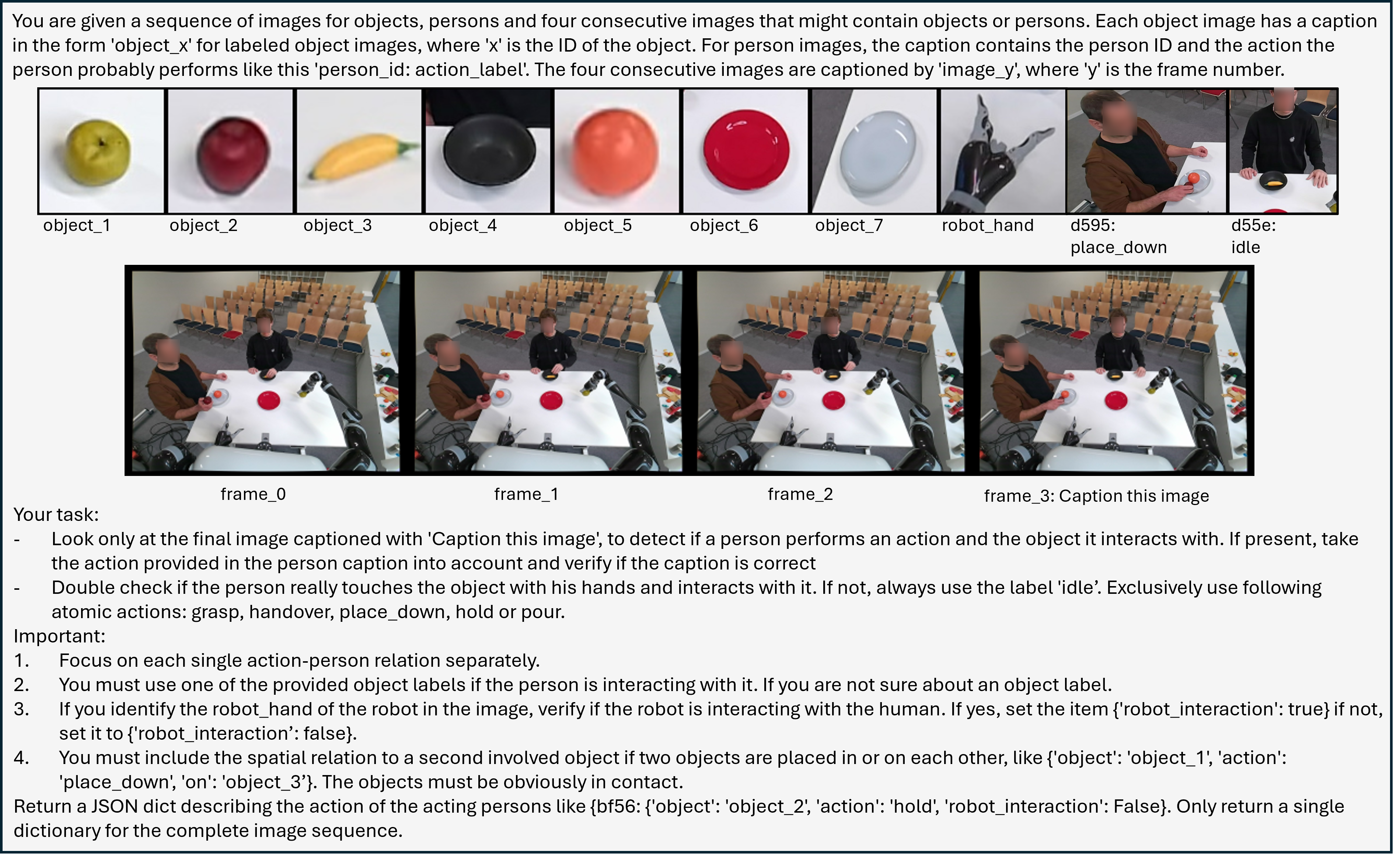}
  \caption{Visualization of the prompt provided to VLM. The prompt begins with a general introduction, followed by cropped object and person images (with person ids), each uniquely identifiable via caption. The robot hand is optionally included to assess interaction. The last four images show the recent captured images before the action trigger. The prompt concludes with a task description guiding the VLM through action inference, object assignment, spatial relation, and robot interaction.}
  \label{figure:MERGE_prompt}
  \vspace{-10pt}
\end{figure*}

In the remainder, we describe how we create such event tuple by identifying physical instances and assign them to specific contextual roles. To this end, we propose five components that structure the scene in a way that offloads low-level perception from the VLM memory, allowing it to focus on higher-level reasoning. These components are: Memory, Object Detector, Person Tracker, Action Detector, and Action Reasoner.
As shown in Figure~\ref{figure:MERGE}, the Object Detector, Person Tracker, and Action Detector extract scene elements without broader context and store them in Memory for consistent instance tracking. The Action Reasoner then builds on these representations to infer context-aware event tuples $\mathcal{T}$. We detail each component below.

\noindent \textbf{Object Detector.} Objects, stored in memory, are not continuously tracked, rather detected using Segment Anything Model (SAM)~\cite{kirillov2023segment} at system start-up to segment the workspace into object candidates. Second, each segmented object $o_j$ is assigned a unique ID and stored in the Memory along with its cropped image: $O = \{o_1, o_2, \ldots\},\text{where }  o_j = \{\text{ID},\ \text{image},\ \text{time}\}$

\noindent \textbf{Person Tracker.} For human actors, temporal consistency is achieved using the body pose tracking functionality provided by the Azure camera SDK \cite{azure_kinect_body_tracking}. The Person Tracker extends the body tracker by assigning a unique ID to each detected actor $a_i$ and forwards cropped person images to the Action Detector. The cropping area in the image is estimated using the 3D body pose projected on the image plane: $A = \{a_1, a_2, \ldots\}, \quad \text{with } \quad a_i = \{\text{ID},\ \text{image},\ \text{time}\}$

\noindent \textbf{Memory}. 
The memory module \(\mathcal{M}\) serves as a central component by maintaining all detected actor instances, object instances, and event tuple within a MongoDB database $\mathcal{M} = \left\{T_k \in \mathcal{T} \right\}$.
Each observed instance is stored as a measurement, enriched with properties such as a unique identifier, cropped image of the instance and a timestamp: $a_i = \{\text{ID},\ \text{image},\ \text{time}\} \text{ and }
o_j = \{\text{ID},\ \text{image},\ \text{time}\}$
This design ensures consistent instance representation across frames and facilitates efficient retrieval of prior observations whenever required for triplet generation.

\noindent \textbf{Action Detector}. 
To detect when an actor interacts with an object, the Action Detection module analyzes the visual input from the Person Tracker to infer an action $x$. For each frame, we obtain a set of actions $X$, each associated with an actor:  
\begin{equation}
X = \{x_1, x_2, \ldots\}, \quad f : A \to X, \quad f(a_i) = x_l ,
\end{equation}
where $A$ is the set of actors and $f$ maps each actor $a_i$ to its predicted action $x_l$. Given a video $V$, spatio-temporal features are extracted using I3D~\cite{carreira2017quo}:  
\begin{equation}
I = \text{conv3d}(V) \in \mathbb{R}^{T \times H \times W \times C}.    
\end{equation}
For each actor $a \in A$, an embedding $r_a$ is obtained via RoI pooling over $I$. In parallel, context features $I_{t,h,w}$ are projected to a reduced representation $E_{t,h,w}$. Actor--context relations are then computed to produce attention maps $A_{a,t,h,w}$, which condition the features as  
\begin{equation}
F_{t,h,w|a} = I_{t,h,w} \odot A_{a,t,h,w},    
\end{equation}
where $\odot$ denotes the elementwise (Hadamard) product. The resulting actor-specific features $F$ highlight context regions relevant to each actor, enabling robust action classification. Given our focus on efficiency, the action detector is intentionally lightweight. The outputs of the Action Detector are finally action labels for each actor independently, which serve two complementary roles: as prior action context that can be incorporated into the VLM prompt, and as trigger signals that, upon the execution of a new action, initiate more refined interpretation by the Action Reasoner.

\begin{figure*}[ht]
    \centering
    \subfloat{\includegraphics[width=0.47\textwidth, height=5.2cm]{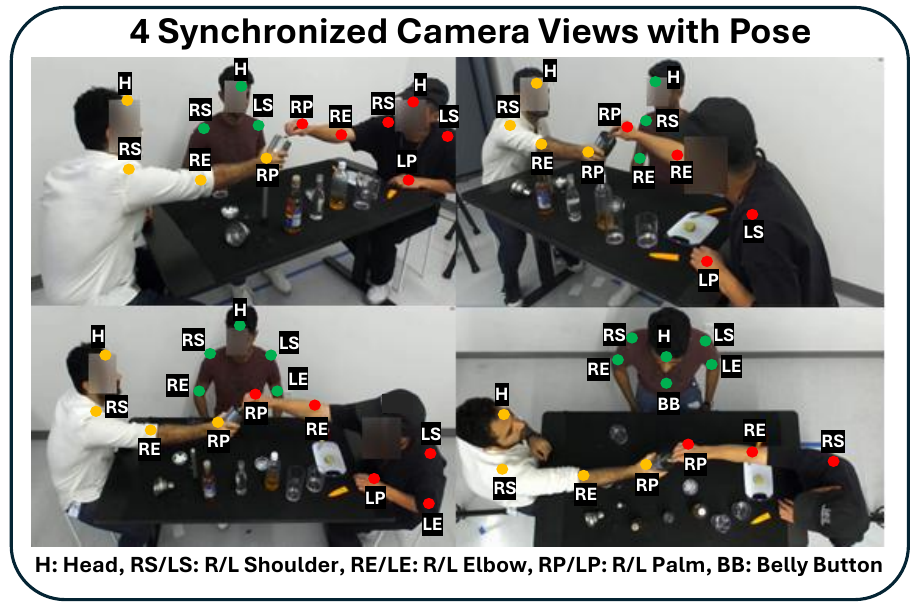}}
    \subfloat{\includegraphics[width=0.47\textwidth, height=5.2cm]{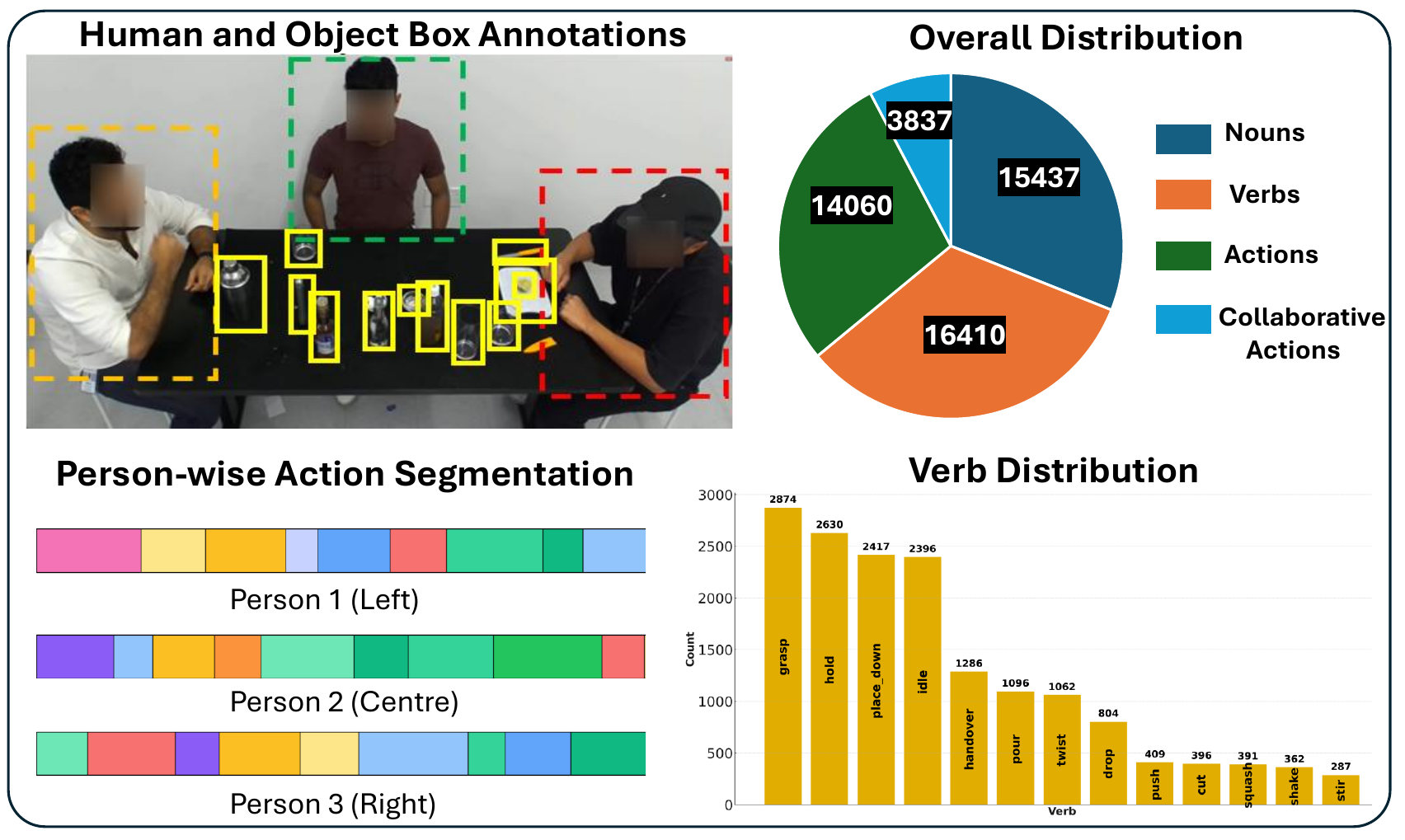}}
    \caption{\textbf{GROUND-Train} provides a rich set of annotations captured from four synchronized camera views. Each video includes person-wise action segmentation labels, 2D pose annotations along with human and object bounding boxes across all views, with the pose and human box annotations further linked through cross-view tracking.}
    \label{fig:GROUND_train}
\end{figure*}


\begin{figure}[ht]
    \centering
    \includegraphics[width=0.45\textwidth, height=3.0cm]{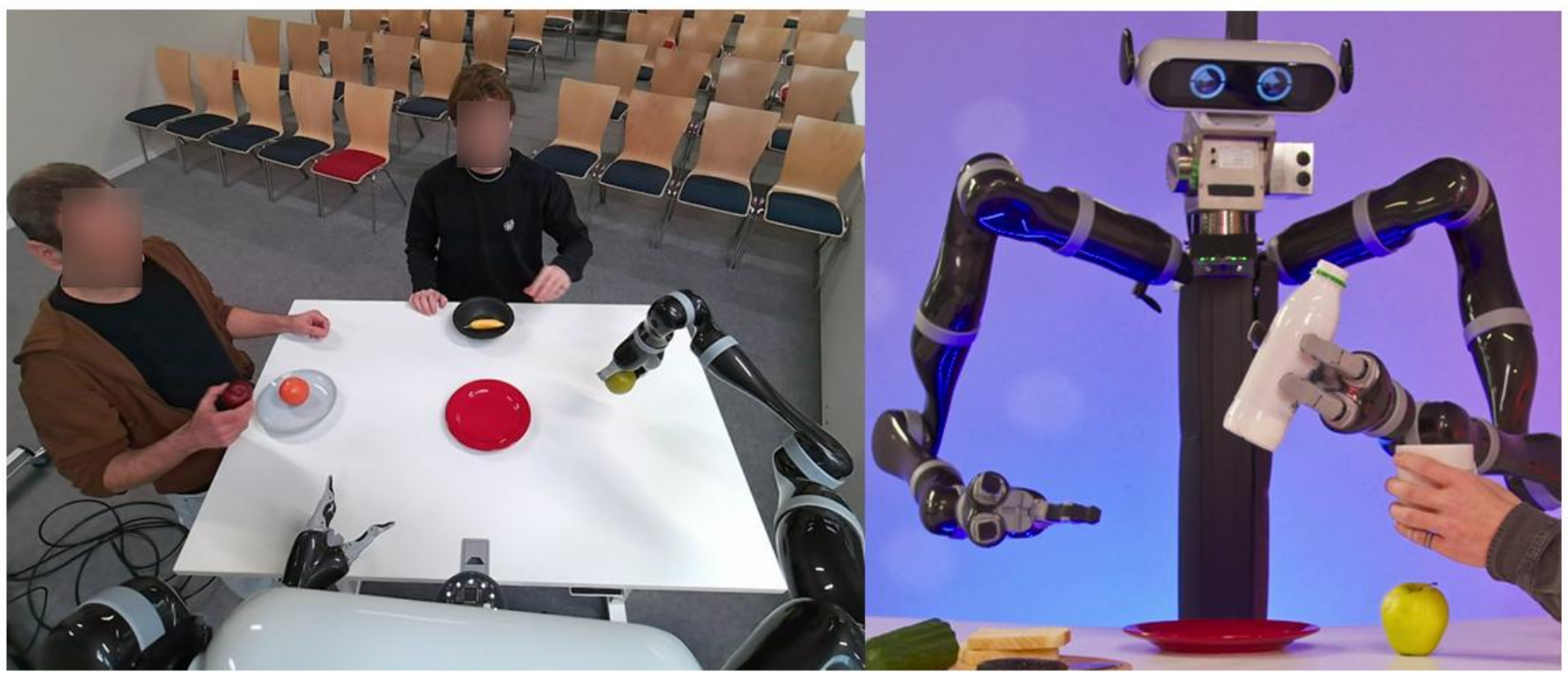} 
    \caption{Left: Example image from \textbf{GROUND-Eval} captured from the robot’s perspective, showing two people and the robot sorting fruits onto two plates. Right: Front-facing view of the robot.}
    \label{fig:GROUND_eval}
    \vspace{-15pt}
\end{figure}

\noindent \textbf{Action Reasoner}. It is the core component (Figure~\ref{figure:MERGE}, bottom) that combines all prior outputs into grounded event tuples $\mathcal{T}$. Built on a VLM, it is guided by the identified actor set $A$ and object instances $O$ to focus on relevant scene elements. For each actor $a_i$ and detected action $x_l$, it infers the involved object $o_j \in O$, the spatial relation $r$, whether a robot interaction with person $i$ occurs, and constructs the tuple $T_i$. Leveraging the VLM’s vision--language capabilities ensures outputs remain grounded in previously identified real-world instances. The final output is the set of tuples per image $\mathcal{T} = \{T_1, T_2, \ldots, T_n\}$.
An event tuple $T = (a, x, o, r, i, t)$ is generated if: the actor $a$ is successfully tracked; an action $x = f(a)$ is detected; an object $o$ used by the actor is identified; a spatial relation $r$ is inferred; a robot interaction $i$ is detected; and a temporal index $t$ is assigned (via action trigger or frame number). A spatial relation \( r \) is constructed by an involved instance \( e \in A \cup O \) and a symbolic relation $\rho$:
\begin{equation}
r = (\rho, e) \quad \text{where} \quad \rho \in \{\text{on},\ \text{in},\ \text{to}\}    
\end{equation}

As shown in Figure~\ref{figure:MERGE_prompt}, the VLM prompt includes: (1) a general instruction; (2) cropped, instance-labeled object and person images (optionally including the robot hand to assess interaction); and (3) the four most recent frames preceding the action trigger. The VLM infers the performed action, involved object, and—if applicable—a second object and its spatial relation to the first object. The final output follows the format: \{'object': 'object\_2', 'action': 'place\_down', 'on': 'object\_4', 'robot\_interaction': false\}

\begin{table*}[]
\caption{Grounding Score (GS) based on comparison with ground truth sequence ($\delta$ = 5s).}
\centering
\setlength{\tabcolsep}{4pt}
\label{table:tsr}
\begin{tabular}{cccccccccccc}
\toprule
\multirow{2}{*}{VLM}            & \multirow{2}{*}{Method}           & \multicolumn{4}{c}{Sorting Fruits} & \multicolumn{2}{c}{Pouring}  & \multicolumn{2}{c}{Handover} & Overall GS     & Overall Runtime (s)      \\ \cline{3-12} 
                                &                                   &  1P   & 2P  & 1P+R & 2P+R          & 2P   & 1P+R                & 2P   & 1P+R                   & $\emptyset$   & $\emptyset$          \\ \hline
\multirow{2}{*}{GPT-4o~\cite{achiam2023gpt}}            & VLM-only  & 0.00 & 0.29 & 0.37 & 0.22 & 0.07 & 0.22 & 0.25 & 0.15 & 0.23 & 0.66    \\
                                                        & Ours      & \textbf{0.46} & \textbf{0.42} & \textbf{0.48} & \textbf{0.47} & \textbf{0.38} & 0.22 & \textbf{0.44} & \textbf{0.50} & \textbf{0.42} & \textbf{0.36} \\ \hline
\multirow{2}{*}{GPT-5~\cite{openaiIntroducingGPT5}}     & VLM-only  & 0.11 & 0.29 & 0.32 & 0.15 & 0.17 & 0.35 & 0.38 & 0.00 & 0.24 & 5.52                \\
                                                        & Ours      & \textbf{0.46} & \textbf{0.36} & \textbf{0.48} & \textbf{0.29} & \textbf{0.42} & \textbf{0.40} & \textbf{0.50} & 0.00 & \textbf{0.38} & \textbf{1.39} \\ \hline
\multirow{2}{*}{Gemini 2.5 Flash~\cite{team2024gemini}} & VLM-only  & 0.00 & 0.19 & 0.13 & 0.07 & 0.07 & 0.07 & 0.13 & 0.04 & 0.10 & 3.23                 \\
                                                        & Ours      & \textbf{0.53} & \textbf{0.43} & \textbf{0.48} & \textbf{0.20} & \textbf{0.22} & \textbf{0.26} & \textbf{0.36} & \textbf{0.36} & \textbf{0.35} & \textbf{0.56} \\ \hline
\multirow{2}{*}{Gemini 2.5 Flash Video~\cite{team2024gemini}} & VLM-only & 0.00 & 0.24 & 0.18 & 0.10 & 0.08 & 0.15 & 0.29 & 0.17 & 0.16 & 2.77                 \\
                                                        & Ours      & \textbf{0.45 }& \textbf{0.43} & \textbf{0.55} & \textbf{0.19} & \textbf{0.45} & \textbf{0.24} & \textbf{0.50} & \textbf{0.22} & \textbf{0.39} & \textbf{0.56} \\ 
\bottomrule                              
\end{tabular}
\vspace{-10pt}
\end{table*}

\section{GROUND Dataset}

To support the development and evaluation of the MERGE system, we introduce a novel dataset, GROUND (\textbf{G}roup \textbf{R}easoning for \textbf{O}bject-centric \textbf{U}nderstanding of \textbf{N}arrative \textbf{D}ynamics). GROUND is designed to facilitate learning and benchmarking in collaborative human–robot interaction scenarios, with a particular focus on fine-grained temporal action segmentation and structured event-level reasoning.
To address objectives, GROUND is divided into two complementary subsets:

\begin{itemize}
\item GROUND-Train: A subset for training and evaluating fine-grained action detection and segmentation.
\item GROUND-Eval: An independently recorded evaluation subset annotated with structured actor–action–object relations for event-level reasoning.
\end{itemize}

Both subsets share a tabletop setup with multiple actors performing individual and collaborative activities, including atomic actions such as \textit{hold}, \textit{pour}, and \textit{handover}. This consistency ensures they complement each other, while differences in recording locations, backgrounds, and tasks provide diversity to demonstrate generalization beyond a single environment. Together, GROUND-Train and GROUND-Eval form a comprehensive benchmark for advancing both low-level perception and high-level reasoning in group interaction scenarios. 

\noindent \textbf{GROUND-Train.}
It comprises 198 unique scenarios, each simultaneously recorded from four distinct camera viewpoints (Figure~\ref{fig:GROUND_train}), yielding 792 synchronized video sequences. The videos capture diverse group configurations—single-person, dyadic, and triadic interactions (1–3 participants)—where individuals prepare drinks following different recipes, requiring both individual actions and coordinated group activities. Each frame is comprehensively annotated with: (a) per-person fine-grained action labels; (b) human bounding boxes and 2D pose estimations; (c) object bounding boxes and semantic categories; and (d) collaborative action labels—\textit{Handover}, \textit{Collaborative Pour}, \textit{Collaborative Twist}, and \textit{Collaborative Drop}. Annotations are consistent across viewpoints, with human boxes and poses linked via cross-view tracking to support multi-view learning and cross-perspective analysis.

The dataset comprises 95 unique action classes (e.g., \textit{hold shaker}, \textit{place\_down glass}, \textit{handover glass} etc), derived from 19 distinct nouns (e.g., \textit{shaker}, \textit{cutting\_board}, \textit{glass}, \textit{muddler} etc) and 13 verbs (e.g., \textit{idle}, \textit{grasp}, \textit{handover}, \textit{cut}, \textit{place\_down}, \textit{drop}, \textit{twist}, \textit{hold}, \textit{pour}, \textit{squash}, \textit{shake}, \textit{push}, \textit{stir}). To the best of our knowledge, this is the first multi-view dataset with diverse annotations—including action segmentation, pose, and bounding boxes—for studying multi-person collaborative instructional activities, offering a valuable benchmark for robotics and computer vision.

\noindent \textbf{GROUND-Eval.}
\label{sec:GROUND_eval}
The dataset is designed to provide fine-grained action reasoning in collaborative human–robot settings and includes detailed situational annotations of multi-actor interactions with robot participation. GROUND-Eval comprises two persons and a robot interacting in tabletop scenarios as shown in Figure~\ref{fig:GROUND_eval}, recorded and annotated in a different environment and setting than GROUND-Train to provide an independent environment for testing. The scenarios are:

\begin{enumerate}
\item \textbf{Sorting Fruits}: A banana, two apples, and an orange are sorted into a bowl or onto a plate. 
\item \textbf{Pouring}: A bottle is used to pour liquid into one of several cups. 
\item \textbf{Handover}: Participants hand various items to each other around the table. 
\end{enumerate}

Each scenario is repeated two times in different constellations:
\begin{itemize} 
\item \textbf{1P}: 1 person performs actions alone. 
\item \textbf{2P}: 2 people perform actions independently or interact. 
\item \textbf{1P+R}: 1 person and the robot perform actions independently or interact. 
\item \textbf{2P+R}: Two people and the robot perform actions independently or interact. 
\end{itemize}

In total, this results in 16 recordings: 8 for sorting fruits, 4 for pouring, and 4 for handovers.
Each video frame is annotated with: a) the full scene image; b) a cropped image of each person instance acting in the scene, labeled by ID; c) sample images of object instances appearing in the scene labeled by ID; d) a full event description consisting of: person ID, action label, if the robot interacts with the acting person, object ID - and if applicable - the spatial relation between two objects participating in an action. 

\section{Experiments} \label{sec:evaluation}

\subsection{Metrics and Evaluation} 
We evaluate MERGE’s ability to generate accurate event tuple in collaborative scenarios. For this purpose, we use the GROUND-Eval dataset (see Section~\ref{sec:GROUND_eval}), which provides detailed annotations for multi-person and human–robot interactions. 
Our evaluation focuses on two aspects: 
(i) the accuracy of situational grounding, i.e. correct identification and ordering of event tuple, and 
(ii) the efficiency of event-driven reasoning compared to state-of-the-art Vision–Language Models (VLMs). 

\noindent \textbf{Grounding Score (GS).}  
We propose a new metric Grounding Score to evaluate our predictions.  
Each ground-truth event is represented as a event tuple  
\( g = (x, o, r, i, t) \) with action \(x\), object \(o\), spatial relation \(r\),  
robot interaction flag \(i\), and start time \(t\).  
Predictions are given by  
\(\hat{g} = (\hat{x}, \hat{o}, \hat{r}, \hat{i}, \hat{t})\).  
The actor is not included in the evaluation measure, as the correct 
actor ID assignment is implicitly encoded by a matching set $T$.
A prediction is considered a match if all fields agree and its predicted time lies within a temporal tolerance \(\delta\) (set to 5s in our experiments), 
\begin{equation}
\mathbb{I}(\hat{g}, g) = 
\mathbb{I}\!\left[\, \hat{x}=x \,\wedge\, \hat{o}=o \,\wedge\, \hat{r}=r \,\wedge\, \hat{i}=i \,\right]
\cdot \mathbb{I}(\hat{t}, t),
\end{equation}
where $\mathbb{I}$ is an indicator function. If multiple predictions fall within this window, we select the one closest in time  
(\(\arg\min |\hat{t}-t|\)). Each prediction can be assigned at most once. Based on this matching, we count true positives (TP), false positives (FP),  
and false negatives (FN). Precision is defined as  
\(P = \tfrac{\mathrm{TP}}{\mathrm{TP}+\mathrm{FP}}\), while recall is defined as  
\(R = \tfrac{\mathrm{TP}}{\mathrm{TP}+\mathrm{FN}}\). The Grounding Score is then 
given as combination of both quantities \(\mathrm{GS} = \tfrac{2PR}{P+R}\). We report GS not only for the overall tuple but also for each of its constituent elements, thereby enabling a more granular analysis.

\subsection{Baselines}

To establish a baseline, we evaluate state-of-the-art VLMs without the structured scene inputs provided by MERGE.
More specifically, we provide the present scene elements about persons and objects as pure textual input along with four consecutive images of the whole scene, so that the VLMs have a similar set-up as MERGE. The VLMs we evaluate are GPT-4o, GPT-5, and Gemini 2.5 Flash, and—leveraging Gemini’s native video capability (similar to Qwen-VL and VideoLLaMA2)—also assess whether providing four consecutive images as video (Gemini 2.5 Flash Video) improves temporal reasoning for action recognition. These widely recognized commercial VLMs represent strong publicly available baselines for comparison with a structured approach like MERGE.

\begin{table*}[]
\caption{Grounding Score (GS) of each role contributing in an event tuple $T$ for each experiment ($\delta$ = 5s).}
\label{table:roles}
\setlength{\tabcolsep}{4pt}
\centering
\begin{tabular}{cccccccccccccccccc}
\toprule
\multirow{2}{*}{VLM}    & \multirow{2}{*}{Method} & \multicolumn{4}{c}{Sorting Fruits} & \multicolumn{4}{c}{Pouring} & \multicolumn{4}{c}{Handover} & \multicolumn{4}{c}{All}     \\ \cline{3-18} 
                        &                                       & $x$ & $o$ & $r$ & $i$ & $x$ & $o$ & $r$ & $i$ & $x$ & $o$ & $r$ & $i$ & $x$ & $o$ & $r$ & $i$ \\ \hline
\multirow{2}{*}{GPT-4o~\cite{achiam2023gpt}}            & VLM-only & 0.32 & 0.49 & 0.33 & 0.58 & 0.19 & 0.25 & 0.27 & 0.30 & 0.20 & 0.36 & 0.48 & 0.48 & 0.26 & 0.41 & 0.34 & 0.49     \\
                                                        & Ours     & \textbf{0.64} & \textbf{0.68} & \textbf{0.49} & \textbf{0.71} & \textbf{0.36} & \textbf{0.56} & \textbf{0.46} & \textbf{0.56} & \textbf{0.46} & \textbf{0.85} & \textbf{0.85} & \textbf{0.85} & \textbf{0.56} & \textbf{0.68} & \textbf{0.53} & \textbf{0.70}   \\ \hline                        
\multirow{2}{*}{GPT-5~\cite{openaiIntroducingGPT5}}     & VLM-only & 0.35 & 0.39 & 0.46 & 0.58 & 0.34 & 0.41 & 0.34 & 0.41 & 0.38 & \textbf{0.69} & \textbf{0.85} & \textbf{0.69} & 0.35 & 0.43 & 0.47 & 0.56         \\
                                                        & Ours     & \textbf{0.52} & \textbf{0.62} & \textbf{0.65} & \textbf{0.83} & \textbf{0.41} & \textbf{0.59} & \textbf{0.47} & \textbf{0.59} & \textbf{0.44} & 0.67 & 0.67 & 0.56 & \textbf{0.49} & \textbf{0.62} & \textbf{0.62} & \textbf{0.76}         \\ \hline
\multirow{2}{*}{Gemini 2.5 Flash~\cite{team2024gemini}} & VLM-only & 0.18 & 0.21 & 0.26 & 0.27 & 0.09 & 0.11 & 0.11 & 0.13 & 0.09 & 0.17 & 0.18 & 0.18 & 0.13 & 0.17 & 0.19 & 0.20        \\
                                                        & Ours     & \textbf{0.55} & \textbf{0.64} & \textbf{0.60} & \textbf{0.69} & \textbf{0.40} & \textbf{0.44} & \textbf{0.52} & \textbf{0.60} & \textbf{0.36} & \textbf{0.64} & \textbf{0.64} & \textbf{0.64} & \textbf{0.50} & \textbf{0.59} & \textbf{0.58} & \textbf{0.66}  \\ \hline
\multirow{2}{*}{Gemini 2.5 Flash Video~\cite{team2024gemini}} & VLM-only & 0.24 & 0.31 & 0.36 & 0.40 & 0.18 & 0.18 & 0.23 & 0.25 & 0.24 & 0.43 & 0.47 & 0.47 & 0.22 & 0.28 & 0.33 & 0.36        \\
                                                        & Ours     & \textbf{0.51} & \textbf{0.59} & \textbf{0.55} & \textbf{0.62} & \textbf{0.50} & \textbf{0.54} & \textbf{0.54} & \textbf{0.61} & \textbf{0.40} & \textbf{0.72} & \textbf{0.72} & \textbf{0.72} & \textbf{0.50} & \textbf{0.59} & \textbf{0.56} & \textbf{0.63}  \\
\bottomrule
\end{tabular}
\vspace{-10pt}
\end{table*}

\begin{table}[t]
\caption{Ablation study on $\delta$ and input image configuration with Overall GS score reported for different VLMs.}
\label{table:abl}
\centering
\setlength{\tabcolsep}{5pt}
\begin{tabular}{ccccc}
\toprule
\multirow{2}{*}{VLM}                                                          & \multirow{2}{*}{Method} & \multicolumn{3}{c}{$\delta$} \\ \cline{3-5} 
                                                                              &                         & $1s$     & $3s$    & $5s$    \\ \hline
\multirow{3}{*}{GPT-4o~\cite{achiam2023gpt}}            & VLM-only            & 0.19                    & 0.22     & 0.23    \\
                                                        & Ours                & 0.28                    & 0.39     & 0.42    \\
                                                        & Ours (cropped)       & \textbf{0.36}           & \textbf{0.52}     & \textbf{0.56} \\ \hline                                                                          
\multirow{3}{*}{GPT-5~\cite{openaiIntroducingGPT5}}     & VLM-only            & 0.20                    & 0.23     & 0.24    \\
                                                        & Ours                & 0.27                    & 0.36     & 0.38    \\
                                                        & Ours (cropped)       & \textbf{0.40}           & \textbf{0.52}     & \textbf{0.53}   \\ \hline
\multirow{3}{*}{Gemini 2.5 Flash~\cite{team2024gemini}} & VLM-only            & 0.08                    & 0.10     & 0.10    \\
                                                        & Ours                & 0.23                    & 0.31     & 0.35    \\
                                                        & Ours (cropped)       & \textbf{0.28}           & \textbf{0.39}     & \textbf{0.44}    \\ \hline
\multirow{3}{*}{Gemini 2.5 Flash Video~\cite{team2024gemini}} & VLM-only            & 0.14                   & 0.16     & 0.16    \\
                                                        & Ours                & 0.22                    & 0.34     & 0.39    \\
                                                        & Ours (cropped)       & \textbf{0.31}           & \textbf{0.43}     & \textbf{0.46}    \\ 
\bottomrule
\end{tabular}
\vspace{-10pt}
\end{table}

\subsection{Results}
\noindent \textbf{Action Detection.} Since object re-identification is handled by VLMs, we train the action detector on verbs only (atomic actions in GROUND-Eval), omitting nouns. Training uses person-associated action labels only, as actions alone trigger events, and excludes additional annotations (e.g., object boxes, 2D poses) for efficiency. Because GROUND-Eval contains only front-view data, we train on the front-view subset of GROUND-Train. The model is trained for 60 epochs on a 165/33 train–val split with batch size 6 and learning rate 0.01, using a single NVIDIA Quadro RTX 6000 GPU. 
We achieve a overall mean average precision (mAP) of 81.8\%, with individual class mAPs as \textit{idle} (98.2\%), \textit{grasp} (76.2\%), \textit{handover} (81.5\%), \textit{cut} (93.8\%), \textit{place\_down} (73.8\%), \textit{drop} (72.4\%), \textit{twist} (90.1\%), \textit{hold} (91.6\%), \textit{pour} (97.7\%), \textit{squash} (91.7\%), \textit{shake} (94.8\%), \textit{push} (19.1\%), \textit{stir} (82.9\%). Out of these 13 verbs, only five are used in GROUND-Eval, i.e. \textit{grasp}, \textit{handover}, \textit{place\_down}, \textit{hold}, \textit{pour}.

\noindent \textbf{Event Grounding.} Table~\ref{table:tsr} presents the GS results for MERGE.
Our analysis reveals that MERGE improves the grounding score by 0.19, which is a factor of around 2 compared to the performance of VLM-only baseline.
This enhanced performance is primarily to be attributed to the trigger signal from the Action Detector, which effectively reduces false detections by selectively guiding event analysis in the image sequence. This can also be observed in Table~\ref{table:roles}, which shows which instances in a tuple were wrongly detected. We can observe that actions $x$ are improved by a GS of around 0.27. Additionally, the structured visual input of cropped objects $o$ improved by 0.29 on average, spatial relations $r$ by 0.24 and the indication if the robot interacts with a person $i$ by 0.29. A comparison between Gemini processing single consecutive images (Gemini 2.5 Flash) and Gemini processing four frames as a video (Gemini 2.5 Flash Video) does not reveal any significant performance improvement. Depending on the scene, one approach may be more advantageous than the other.

Beyond performance gains, we also report per-frame runtime in Table~\ref{table:tsr}. MERGE achieves a 4$\times$ reduction in computation time, running at an average of 0.77s (including $\sim$20 ms for action detection) compared to 3.14s for VLM-only approaches. This efficiency stems from the perception pipeline, where the VLM is invoked only when a trigger signal is raised. There is a one-time object detection computation cost of $\sim$0.5 s at the beginning of each video, but we exclude it from the runtime comparison since the VLM baselines also rely on ground-truth object label inputs. Even if this cost were included, its impact would be marginal when averaged over all frames.


\noindent \textbf{Ablation Study.} For a fair comparison with baselines, we provided MERGE with full scene images during evaluation.
However, in a real robotic deployment, MERGE can rely solely on cropped person images obtained from the person tracker.
As shown in Table~\ref{table:abl}, this leads to an additional Grounding Score (GS) improvement of 0.13 over the non-cropped version, and 0.35 compared to plain VLMs — emphasizing the benefit of structured, instance-focused inputs.
Table~\ref{table:abl} also illustrates how GS varies with the temporal tolerance $\delta$.
As expected, $\delta$=1s results in a clear performance drop across all methods, highlighting that this threshold is too strict for real-world settings, where minor misalignments between perception and annotation are common.
In contrast, $\delta$=3 and $\delta$=5s produce similar and robust results, offering a reasonable trade-off between temporal precision and robustness in human–robot interaction.

\subsection{Discussion and Limitations}
Our evaluation shows that MERGE moves beyond class-level recognition to grounded, instance-level identification—essential for multi-party scenarios with overlapping roles, object reuse, and temporal dependencies. Its VLM-agnostic, event-driven design improves accuracy and efficiency, though selective triggering introduces a trade-off between efficiency and recall. A key limitation is that object detection runs only at initialization and is not updated; future work could enable continuous memory–scene comparison to capture new objects, though instance-level disambiguation remains challenging. Additionally, while GROUND-Train provides rich annotations, we use only a subset for GROUND-Eval, leaving room for broader dataset utilization, especially for action segmentation~\cite{ghoddoosian2022weakly, ghoddoosian2023weakly}. Overall, this work motivates finer pipeline selectivity, stronger VLM resolution, and richer data use toward adaptive, collaborative robots.

\section{Conclusion}
This work introduced MERGE, a framework that leverages a lightweight perception module to guide VLMs to maintain consistent actor–object representations and generate structured event tuples in dynamic group settings. We also developed GROUND, a dataset of multi-human–robot collaborations with detailed role-aware annotations and new evaluation metrics. Experiments show that MERGE achieves both higher accuracy and faster runtime than strong VLM baselines, underscoring the promise of event-driven perception for group-aware HRI. 
Looking ahead, MERGE and GROUND provide a foundation for more adaptive, memory-driven robots that can collaborate effectively in real-world multi-actor environments.

\bibliographystyle{IEEEtran}
\bibliography{IEEEfull}

\end{document}